\documentclass[11pt]{article}

\usepackage{acl2023}

\usepackage{times}
\usepackage{latexsym}
\usepackage[pdftex]{graphicx}

\usepackage[T1]{fontenc}

\usepackage[utf8]{inputenc}

\usepackage{microtype}

\usepackage{inconsolata}

\def\Underline{\setbox0\hbox\bgroup\let\\\endUnderline}
\def\endUnderline{\vphantom{y}\egroup\smash{\underline{\box0}}\\}
\def\|{\verb|}
\usepackage{breakurl}

\title{Training Generative Question-Answering on Synthetic Data Obtained from an Instruct-tuned Model}

\author{Kosuke Takahashi, Takahiro Omi, Kosuke Arima\\
        Stockmark \\ kosuke.takahashi, takahiro.omi, kosuke.arima@stockmark.co.jp \AND
  Tatsuya Ishigaki \\
  \normalsize{National Institute of Advanced Industrial Science and Technology} \\ishigaki.tatsuya@aist.go.jp}

\begin{document}
\maketitle
\begin{abstract}
This paper presents a simple and cost-effective method for synthesizing data to train question-answering systems.
For training, fine-tuning GPT models is a common practice in resource-rich languages like English, however, it becomes challenging for non-English languages due to the scarcity of sufficient question-answer (QA) pairs.
Existing approaches use question and answer generators trained on human-authored QA pairs, which involves substantial human expenses.
In contrast, we use an instruct-tuned model to generate QA pairs in a zero-shot or few-shot manner.
We conduct experiments to compare various strategies for obtaining QA pairs from the instruct-tuned model.
The results demonstrate that a model trained on our proposed synthetic data achieves comparable performance to a model trained on manually curated datasets, without incurring human costs.
\end{abstract}

\section{Introduction}

Fine-tuning large language models (LLMs) has been proven effective for enhancing question-answering systems~\cite{NEURIPS2019_c20bb2d9}.
However, extending this approach to languages other than English presents challenges due to the scarcity of adequate QA pairs for training.
In this study, we specifically target Japanese as a representative non-English language.
We propose a straightforward approach that synthesizes Japanese QA pairs using an instruct-tuned model.\footnote{Our experiments utilize OpenAI's ChatAPI with the {\it{gpt-3.5-turbo-0613}} model.}

\begin{figure}[t]
\begin{center} 
\includegraphics[width=7.6cm]{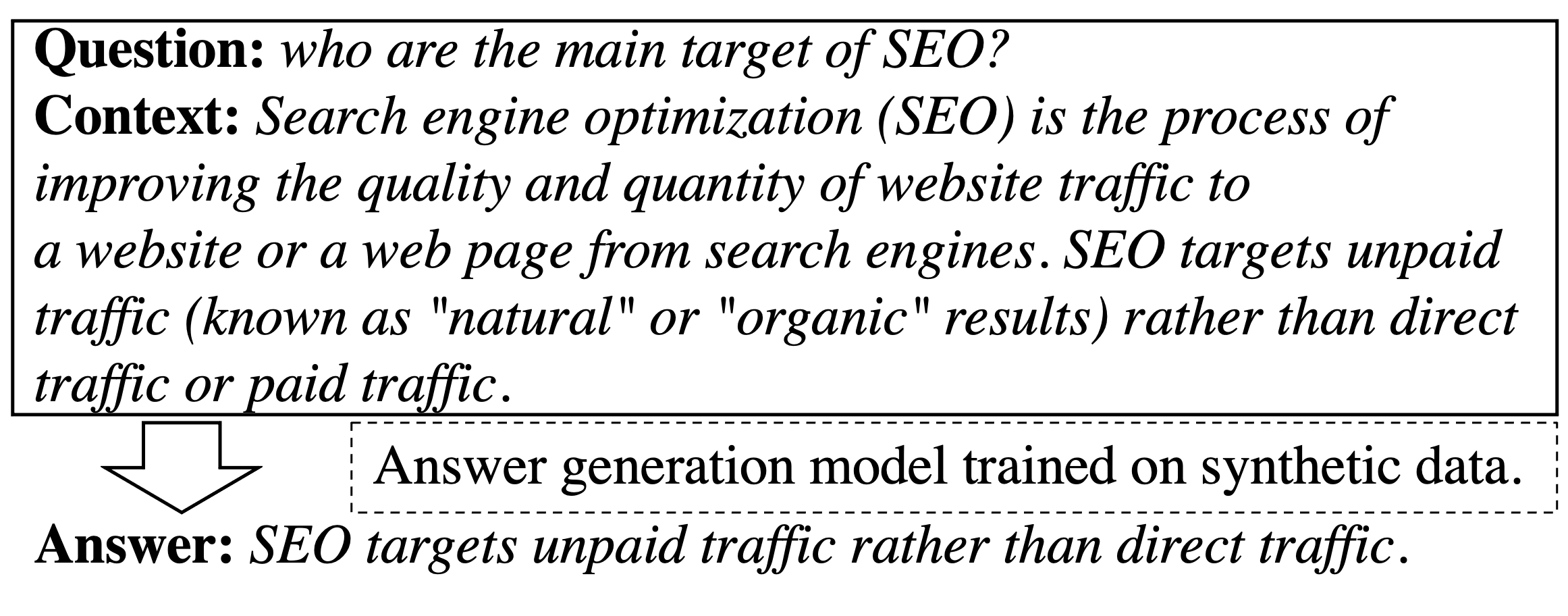}
\vspace{-0.1cm}
\caption{The task of the generative context-aware QA.}
\label{fig:task}
\end{center}
\end{figure}

Question-answering tasks can be categorized into two main settings: questions with context and without context~\cite{JGLUE}.
In this study, we focus on the context-based setting as shown in Figure \ref{fig:task}.
In this setting, the system takes a question along with the accompanying context as input.
The model generates an answer by utilizing the information provided within the context.
On the other hand, the setting without context involves the system processing only the question as input.

\vspace{-0.1cm}

We present a straightforward yet cost-effective method for generating synthetic question-answer (QA) pairs.
Existing QA systems are trained on either human-authored datasets or automatically generated QA pairs~ \cite{sachan-xing-2018-self,tang-etal-2018-learning}, both leading to high labor costs.
By contrast, this paper investigates utilizing an instruct-tuned model inspired by their reasonable ability to produce synthetic dataset~\cite{ChatGPT-outperforms}.
We use a context as input and generate both the corresponding question and its answer.
The instruct-tuned model allows us to produce QA pairs in a zero-shot or few-shot manner, eliminating the need for manual curation.

Our experiments compare question-answering systems fine-tuned on synthetic data generated through various strategies.
Specifically, we explore different sources of contexts, the number of shots fed into the instruct-tuned model, and the quantity of QA pairs generated.
The evaluation on JSQuAD's evaluation dataset~\cite{JGLUE} provides three findings.
Firstly, employing contexts extracted from a corpus with similar characteristics to the evaluation dataset yields improved performance.
Secondly, the one-shot strategy outperforms the zero-shot approach.
Lastly, generating three QA pairs for each context is more effective than generating a lower number of QA pairs.
The top-performing model fine-tuned on our synthetic data exhibits comparable performance to models trained on manually curated data.




\section{Related Work}

Existing QA focus on two major settings: "closedQA" with context and "commonsens-QA" without context~\cite{JGLUE}.
For the former, which we target, the QA systems receive a question along with a context, such as a Wikipedia article, and generate an answer.
On the other hand, in the latter setting, the systems only receive a question as input.

There are two types of QA systems: extractive and generative.
Extractive methods extract an answer as it is from the context by models like BERT~\cite{rajpurkar-etal-2016-squad}, while generative methods often use the expressions that are not in the context by models like T5~\cite{raffel-t5} or GPT~\cite{NEURIPS2020_1457c0d6}.
Our focus is on the latter.


While several manually created datasets exist in English, such as SQuAD \cite{rajpurkar-etal-2016-squad} and QuALITY \cite{pang-etal-2022-quality}, these resources do not directly apply to the Japanese language.
For Japanese, JSQuAD~\cite{JGLUE} and JAQKET\footnote{\url{https://www.nlp.ecei.tohoku.ac.jp/projects/jaqket/\#Reference}} are available.
We use JSQuAD\footnote{Strictly, JSQuAD is not for evaluating generative QA, but the span extraction-based setting. We use this data because there is no common evaluation data in Japanese for generative QA. Our models generate answers not extract spans, thus, we also conduct human evaluations.} because the evaluation data of JAQKET is not public.

Existing studies synthesize QA pairs by two main approaches: supervised~\cite{lee-etal-2020-generating,sachan-xing-2018-self,tang-etal-2018-learning} and unsupervised~\cite{puri-etal-2020-training}.
The supervised approaches train question-answer generators using manually created datasets.
Our approach generates QA pairs from contexts in a zero-shot or few-shot manner, eliminating the need to train generators.
In the unsupervised approach, \citet{puri-etal-2020-training} uses a named entity recognizer (NER) for answer candidate extraction while our approach uses only an instruct-tuned model in end-to-end and does not require NER.

\section{Synthesizing QA Pairs}
\label{sec:method}

\begin{figure}[t]
\setbox0\vbox{
\vbox{Based on the given texts, please make a pair of answerable question and answer.\\
Please make the answer in Japanese polite language. \\
Please respond in the JSON format.\\
\\
\|## example|\\
texts:"texts to extract the pair of question and answer"\\
output:\|{|"Question":"the question that can be answered from the texts", "Answer":"the answer to the question"\}\\
\\
\|## input|\\
texts:\|{QA context}|\\
output:
}
}
\centerline{\fbox{\box0}}
\caption{An example of zero-shot prompt to generate a pair of QA.}
\label{fig:prompt_zero-shot}
\end{figure}

\begin{figure}[t]
\setbox0\vbox{
\vbox{texts:"Resolving technical debt is difficult; we look at JAL's challenge...(omitted)...JAL's watchword is Go To Cloud...(omitted),\\\\
output:\|{|"Question":"What watchwords does Japan Airlines stand for?", "Answer":"JAL's watchword is Go To Cloud."\}
}
}
\centerline{\fbox{\box0}}
\caption{An translated sample of the ``\#\# example'' part in one-shot prompt. Note that the original is in Japanese.}
\label{fig:prompt_example_one-shot}
\end{figure}

We describe our approach in this section.

\subsection{Source Contexts and Filtering}
\label{sec:context}

We generate $N$ question-answer pairs from each context.
$N$ is set to one or three in our experiments.
We compare three specific sources of contexts: 1) a random sample of 6,000 Japanese Wikipedia articles (\texttt{wiki}), 2) a random sample of 6,000 news articles (\texttt{news}), and 3) contexts in JSQuAD's training dataset (\texttt{JSQuAD}).
To collect the news articles, we gathered the most accessed articles from a search engine~\footnote{The URL of the engine/dataset is hidden to preserve the anonymity of authors, and will be shown after acceptance} during the period from May 2022 to May 2023.
We limit each context to the first 300 characters before generating QA pairs by the instruct-tuned model.

\subsection{Prompts for Generating QA Pairs}
We provide examples of zero-shot and one-shot prompts with the setting $N=1$ in Figure \ref{fig:prompt_zero-shot} and Figure \ref{fig:prompt_example_one-shot}, respectively.
These prompts aim to generate QA pairs from a context.
In the zero-shot prompt, we first present the task instructions, followed by an explanation of the structure oh how an input text is represented, and their desired output JSON structure as shown in the ``\#\# example'' section.
For the setting $N>1$, we modify the example of the JSON structure to include more QA pairs.
Then, we write an input text in the ``\#\# input'' section.
In the zero-shot prompt setting, we only write the format of input and output structures, without including actual texts or the expected question-answer pairs corresponding to the context.
On the other hand, in the one-shot prompt, we replace the ``\#\# example'' section in \ref{fig:prompt_zero-shot} with the prompt shown in Figure \ref{fig:prompt_example_one-shot}.
Unlike the zero-shot prompt, the one-shot prompt includes actual example contexts and their corresponding expected QA pairs.
To better understand the effects of prompt engineering, we compare these two prompts in our experiments.
The tuples of a context and generated QA pairs are used to fine-tune a GPT by the prompt shown in Figure \ref{fig:prompt_calm}.






\section{Experiments}
\label{sec:experiments}

\noindent
\textbf{Evaluation Dataset and Compared Models: }
We use the JSQuAD~\cite{JGLUE} for evaluation.
This evaluation data contains 4,470 human-authored QA pairs given Wikipedia articles as contexts.
We use whole evaluation data for the automatic evaluation while randomly sampled 500 instances are used for manual evaluation.

We conduct a comprehensive comparison by exploring various combinations of contexts, the number of generated QA pairs denoted as $N$ and prompts.
Regarding contexts, we consider three options: \texttt{wiki}, \texttt{news}, \texttt{JSQuAD}, and, as detailed in Sec. \ref{sec:context}.
For $N$, we compare $N=1$ and $N=3$.
We compare zero-shot and one-shot prompts~\footnote{We are constrained to one-shot due to the input length limit of ChatGPT.}.

Our proposed models are compared with two models: 1) a plain GPT model without fine-tuning and 2) a model fine-tuned on QA pairs from the JSQuAD training dataset (\texttt{Human}), where these QA pairs are human-authored while our proposed QA pairs are not human-authored.

\paragraph{Fine-tuning}
We use the synthesized QA pairs to fine-tune the Japanese version of GPT-NeoX~\cite{black-etal-2022-gpt}\footnote{\url{https://huggingface.co/cyberagent/open-calm-7b}}.
To achieve improved speed, we employ LoRA fine-tuning \cite{hu2022lora}.
In generating answers, we use a prompt in the zero-shot setting (Figure \ref{fig:prompt_calm}).

\begin{figure}[t]
\setbox0\vbox{
\vbox{\#\# Instruction
\newline\{QUESTION\}
\newline
\newline\#\# Context
\newline\{CONTEXT\}
\newline
\newline\#\# Response}
}
\centerline{\fbox{\box0}}
\caption{The prompt to generate answers with the fine-tuned GPT-NeoX.}
\label{fig:prompt_calm}
\end{figure}



\noindent
\textbf{Metrics: }
For automatic evaluation, we employ BERTScore~\cite{BERTscore} and BLEU~\cite{papineni-etal-2002-bleu}.
BERTScore is implemented on our own with a Japanese BERT model.\footnote{\url{https://huggingface.co/cl-tohoku/bert-base-japanese-v3}} 
As for BLEU, SacreBLEU library~\cite{sacrebleu} is used.

These automatic metrics may not directly capture the correctness of an answer to a given question.
To address this, we also conduct manual evaluations by human judges.
We ask four judges, who are experts in natural language processing or linguistics, to assess whether the generated answer is correct or not.
We showed tuples of questions, answers, and contexts to the judges.
We report the accuracy obtained from the manual evaluation.


\noindent
\textbf{Parameters}
We conducted a grid search for tuning parameters: batch size, learning rate, the number of epochs, as well as LoRA's hyperparameters (specifically $\alpha$ and $r$).
The range of values explored during this search is provided in Table \ref{tab:hyper_parameres}. 
Subsequently, the model that attained the highest BERTScore was chosen for evaluation.

\begin{table}[tb]
    \small
    \centering
    \begin{tabular}{l}\hline
         Batch Size: \{4, 8\},\\ Learning Rate: \{0.00001, 0.00005, 0.000001\},\\
         Epoch: \{3, 4, 5,\}, $r$: \{4, 8, 16, 64, 128\}, 
         $\alpha$: \{1, 4, 16\} \\\hline
    \end{tabular}
    \caption{\label{tab:hyper_parameres}The search range values in LoRA fine-tuning.}
\end{table}

\section{Results}

In this section, we present the results on JSQuAD.

\subsection{Automatic Evaluation}

Our primary interest lies in examining the impact of each strategy for synthesizing QA pairs on the performance of the downstream question answering task. 
Specifically, we focus on comparisons involving different contexts, prompts, and the quantities of automatically generated QA pairs.

Table \ref{tab:metrics_results} presents the scores of BERTScore and BLEU obtained by varying the contexts while keeping other settings, i.e., $N$ and prompts are fixed.
The table is divided into five sections.
Starting from the top, the first section displays scores for QA models trained on human-authored QA pairs (\texttt{Human}) from the JSQuAD training dataset, along with the plain GPT model (\texttt{GPT}) without fine-tuning.
The second and third sections showcase scores obtained when $N$ is fixed to one, but we vary the prompts to zero-shot and one-shot.
The fourth and fifth sections represent scores when we use $N=3$.

\noindent
\textbf{Impact of Context on Performance: }
We observe that using contexts extracted from the \texttt{news} dataset yields relatively low scores, e.g., 0.713 and 0.747 in terms of BERTScore for zero-shot and one-shot settings with $N=3$, respectively.
The \texttt{wiki} context performs better (0.706 and 0.838) than \texttt{news} (0.713 and 0.747) for the same settings.
Notably, the \texttt{JSQuAD} context achieves the highest BERTScore of 0.863 and 0.889 with $N=1$ and $N=3$, respectively.
The results suggest that using Wikipedia as context provides an advantage, likely because the JSQuAD evaluation data is also derived from Wikipedia.

\noindent
\textbf{Impact of Prompts on Performance: }
The one-shot prompt is more effective.
As shown in Table \ref{tab:metrics_results}, the model fine-tuned on the zero-shot QA pairs ($N=1$) generated from the contexts in \texttt{JSQuAD} training dataset achieves a BERTScore of 0.724.
However, the one-shot prompts with $N=1$ exhibit a significant performance gain, reaching a BERTScore of 0.863.

\noindent
\textbf{Effect of the Number of Generated QA Pairs on Performance: }
As we increase the number of QA pairs for context, there is a gain of 2.6 points in BERTScore (from 0.863 to 0.889).
Remarkably, the achieved BERTScore of 0.889 is comparable to that of a model trained on human-authored QA pairs (0.899), despite our approach not utilizing any human-authored QA pairs.

\begin{table}[tb]
    \small
    \centering
    \begin{tabular}{c|c|c|c|c}
         context & $N$ & prompt &  BERTscore & BLEU\\\hline\hline
         \texttt{Human} & - & - & 0.899 & 5.64\\
         GPT & - & - & 0.601 & 0.00\\\hline
         \hline         
         news & 1 & zero & 0.697 & 0.02 \\
         wiki & 1 & zero & 0.713 & 0.03\\
         JSQuAD & 1 & zero & 0.724 & 1.55 \\\hline
         news & 1 & one & 0.738 & 0.11 \\
         wiki & 1 &  one & 0.775 & 0.09\\
         JSQuAD & 1 &  one & 0.863 & 4.83\\ \hline
         news & 3 & zero & 0.713 & 0.38\\
         wiki & 3 & zero & 0.706 & 0.23\\
         JSQuAD & 3 &  zero & 0.740 & 1.85\\\hline
         news & 3 & one & 0.747 & 1.25\\
         wiki & 3 &  one & 0.838 & 1.66\\
         JSQuAD & 3 &  one & \bf{0.889} & \bf{6.77}\\ \hline
    \end{tabular}
     \caption{\label{tab:metrics_results}Performances on different contexts and numbers of generated QA pairs.}

\end{table}


\begin{table}[t]
    \small
    \centering
    \vspace{0.15cm}
    \begin{tabular}{c|c}\hline
         QA Pairs &  Accuracy (\%)\\\hline\hline
        \texttt{JSQuAD ($N=3$, one-shot prompt)} & \bf{45.4} \\
        \texttt{Human} & 38.4 \\\hline
        Gold & 90.4\\\hline
    \end{tabular}
    \caption{\label{tab:human_evaluation}Accuracy calculated as the number of correct question-context-answer tuples divided by the total 500 evaluation instances.}

\end{table}

\subsection{Evaluation by Human Judges: }
We present the results of the manual evaluation.
Table \ref{tab:human_evaluation} shows the comparisons between three outputs: answers generated by 1) our best performing model (\texttt{JSQuAD} ($N=3$), and one-shot prompt) and 2) a model that is fine-tuned on human-authored QA pairs from the JSQuAD training dataset, and 3) gold answers in JSQuAD evaluation dataset.
Remarkably, despite our approach does not use any human-authored QA pairs, the achieved accuracy is 45.4\% while the model fine-tuned on human-authored QA pairs achieves only 38.4\% in terms of accuracy.
\citet{ChatGPT-outperforms} mention that automatic annotation with an instructor-tuning model has higher quality than annotations by crowd-workers, and our results are consistent with their claim.
Note that the performance of both fine-tuned models falls significantly behind the Gold standard (90.4\%), indicating ample room for improvement.


\section{Conclusions}

This paper proposed to use an instruction-tuned model for synthesizing QA pairs.
Our experimental results demonstrate that the models trained on automatically generated QA pairs achieve comparable or even superior performance compared to the fine-tuned model trained on human-authored QA pairs.
In future studies, we plan to explore the relationship between the diversity of automatically generated QA pairs and their impact on the performance of downstream QA tasks.

\bibliography{anthology,custom}
\bibliographystyle{acl_natbib}

\label{sec:appendix}

\end{document}